\documentclass{article}
    \usepackage{spconf,amsmath,graphicx,multirow}
    \usepackage{makecell}

    \title{IMAGE AESTHETICS ASSESSMENT USING COMPOSITE FEATURES FROM OFF-THE-SHELF DEEP MODELS}
    \name{Xin Fu, Jia Yan, Cien Fan}
    \address{Electronic Information School, Wuhan University, China}
\begin{document}
    %
    \maketitle
    \begin{abstract}
    Deep convolutional neural networks have recently achieved great success on image aesthetics assessment task. In this paper, we propose an efficient method which takes the global, local and scene-aware information of images into consideration and exploits the composite features extracted from corresponding pretrained deep learning models to classify the derived features with support vector machine. Contrary to popular methods that require fine-tuning or training a new model from scratch, our training-free method directly takes the deep features generated by off-the-shelf models for image classification and scene recognition. Also, we analyzed the factors that could influence the performance from two aspects: the architecture of the deep neural network and the contribution of local and scene-aware information. It turns out that deep residual network could produce more aesthetics-aware image representation and composite features lead to the improvement of overall performance. Experiments on common large-scale aesthetics assessment benchmarks demonstrate that our method outperforms the state-of-the-art results in photo aesthetics assessment.
    \end{abstract}
    \section{Introduction}
    \label{sec:intro}
    
    Photographic devices like digital camera and smartphone being widely spread allow individuals to take photos more conveniently than ever. Meanwhile, great effort and time must be paid to sift through the piles of images stored in devices and cloud storage. Therefore, automatically picking out aesthetically pleasing images is very useful under such circumstances. Figure 1 shows us some examples of good and bad aesthetic images.
    
    Recent years, the research community addressed this challenging problem by developing ways to classify the images into binary categories of high quality and low quality \cite{DengImageAestheticAssessment2017}. Early work mainly focused on various hand-crafted aesthetic features and feature representations such as SIFT or color descriptors \cite{MurrayAVAlargescaledatabase2012}. With the evolution of deep learning, deep neural network (DNN), especially convolutional neural network (CNN), has been successfully used in various fields, such as image classification \cite{KrizhevskyImagenetclassificationdeep2012,SimonyanVeryDeepConvolutional2014,HeDeepResidualLearning2016}, object detection, scene recognition \cite{ZhouPlaces10million2017} and so on. The state-of-the-art performance achieved by deep neural network proves its powerful ability of feature representation for various visual tasks. Consequently, deep learning-based techniques have been adopted in aesthetics assessment tasks over the past few years and have successfully gained better performance than conventional approaches \cite{DengImageAestheticAssessment2017,MurrayAVAlargescaledatabase2012, LuRAPIDRatingPictorial2014}. 
    
    \begin{figure}[!tbp]
    \begin{minipage}[b]{.48\linewidth}
      \centering
      \centerline{\includegraphics[width=3.2cm]{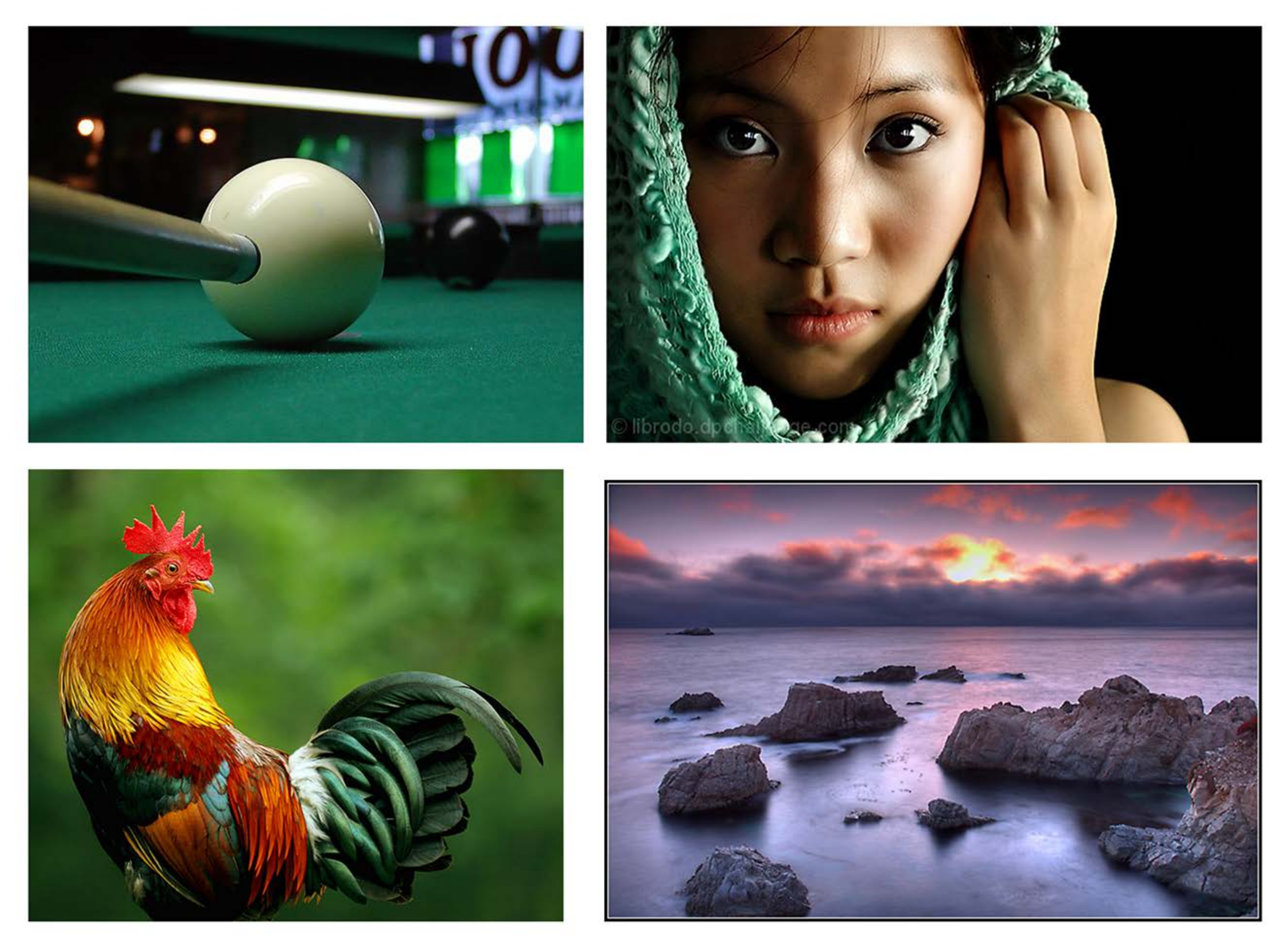}}
      \centerline{(a)}\medskip
    \end{minipage}
    \hfill
    \begin{minipage}[b]{0.48\linewidth}
      \centering
      \centerline{\includegraphics[width=3.2cm]{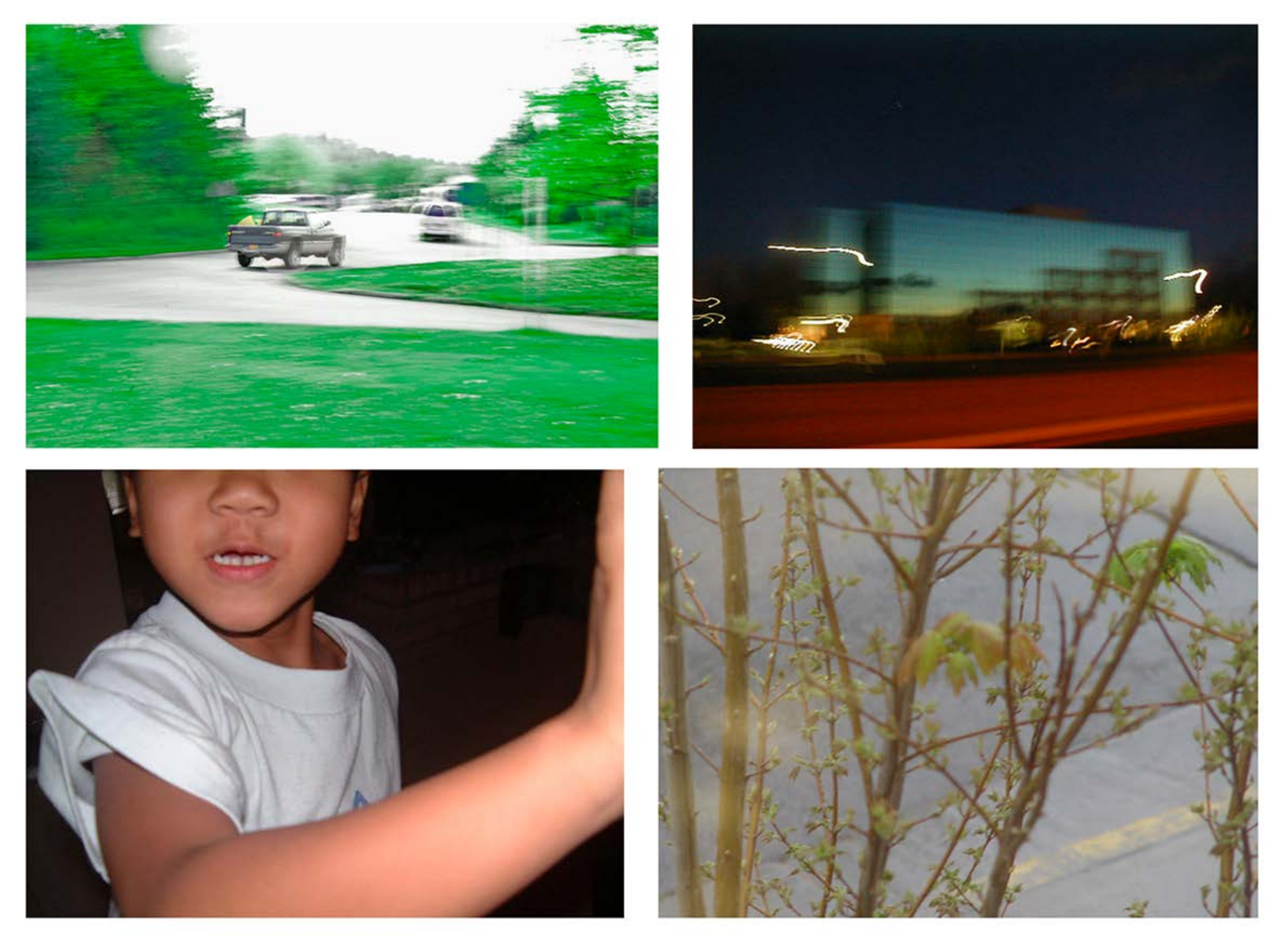}}
      \centerline{(b)}\medskip
    \end{minipage}
    \caption{Examples of good and bad aesthetic images from AVA \cite{MurrayAVAlargescaledatabase2012} dataset. (a) good aesthetic images, (b) bad aesthetic images.}
    \label{fig:figure1}
    \end{figure}

    Very deep convolutional networks have been central to the largest advances in image recognition performance in recent years. Existing popular deep neural networks are carefully designed for visual tasks and have been trained on large-scale datasets comprised of millions of images like ImageNet \cite{DengImageNetlargescalehierarchical2009}, so these deep neural networks have powerful ability of extracting generic image representation that could be applied to other similar visual tasks \cite{RazavianCNNFeaturesofftheshelf2014,DonahueDeCAFDeepConvolutional2014}. The generated generic features contain information with respect to their aesthetics as well. Therefore, to achieve better results, more advanced model is needed. However, shallow neural networks are still widely adopted in many deep learning-based techniques for aesthetics assessment. Furthermore, nearly all of these methods require fine-tuning or training a model from scratch, which is so inefficient that the process typically consumes days or weeks \cite{JinEfficientDeepAesthetic2016}. Our work tries to overcome these disadvantages by making the most of the potential of existing trained deep learning models.
    
    \begin{figure*}[ht]
        \centering
        \includegraphics[width=0.65\textwidth]{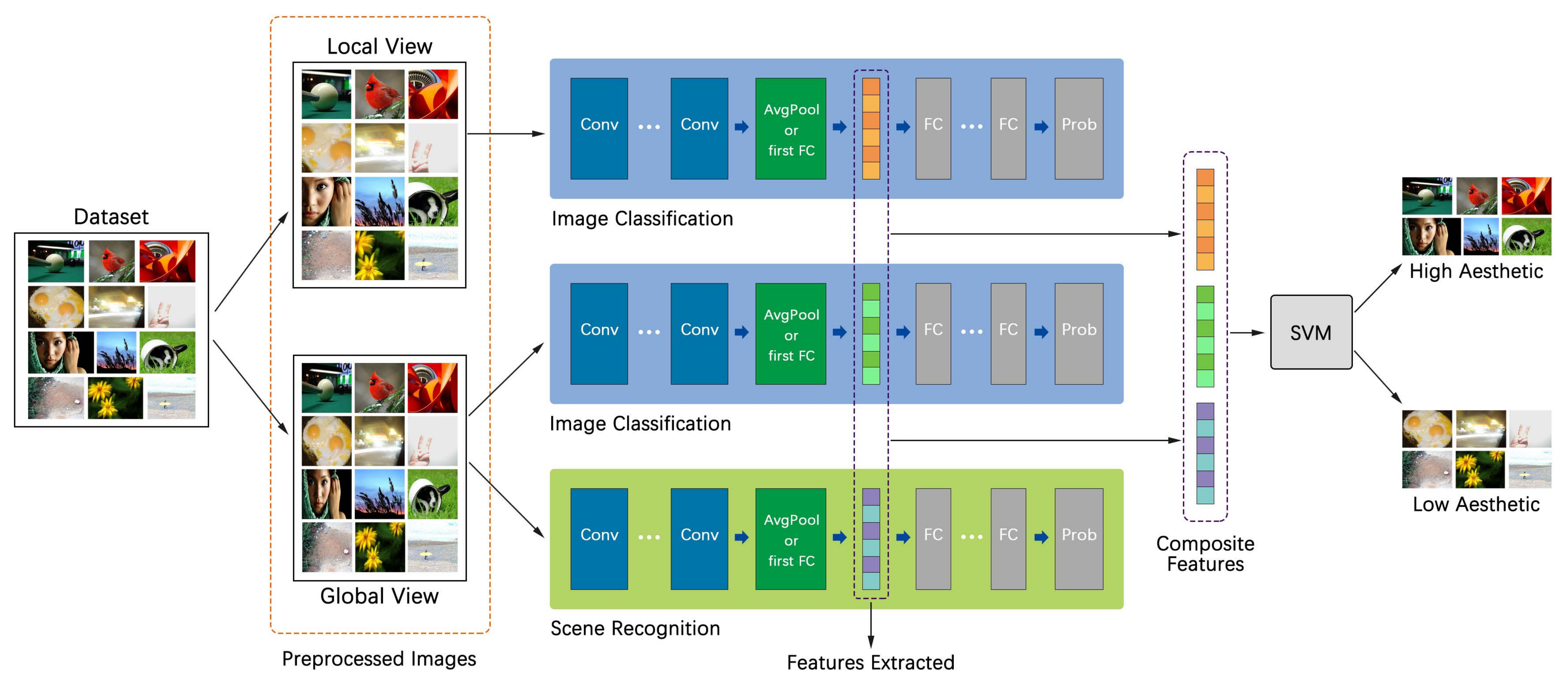}
    \caption{The architecture of our proposed method.}
    \label{fig:figure2}
    \end{figure*}
    
    In this paper, we propose a more efficient method using off-the-shelf deep neural networks for both image classification and scene recognition to extract deep representation of images from three perspectives. By bringing the global, local and scene-aware representations together to yield the composite features intended for further classification, we are able to enhance the results of aesthetics assessment. Compared to other state-of-the-art approaches that need re-training or fine-tuning a DNN model, our proposed neural network training-free method have achieved better performance.
    
    \section{Related Work}
    In this section, we will give a brief overview of recent works from two different aspects.

    \subsection{Image Aesthetics Assessment}
    Methods in image aesthetics assessment could generally be divided into three distinct categories: classical handcrafted low-level features, generic features based on image descriptors, and the contemporary approach of utilizing deep learning models.
    Datta et al. \cite{DattaStudyingAestheticsPhotographic2006} proposed visual features based on standard photography and visual design rules to encapsulate aesthetic attributes from low-level image features. Marchesotti et al. \cite{MarchesottiDiscoveringBeautifulAttributes2015} proposed to learn aesthetic attributes from textual comments on the photographs using generic image features. 
    
    Recently, deep learning methods have been applied to image aesthetic assessment \cite{JinEfficientDeepAesthetic2016,LuRAPIDRatingPictorial2014,MaiCompositionPreservingDeepPhoto2016} and have significantly improved the prediction precision against previous non-deep methods. Tian et al. \cite{TianQueryDependentAestheticModel2015} proposed a query-dependent aesthetic model based on feature representation learned from CNN. Dong et al. \cite{DongPhotoQualityAssessment2015} proposed to adopt the generic features from the penultimate layer output of AlexNet with spatial pyramid pooling. Wang et al. \cite{Wangmultiscenedeeplearning2016} proposed a CNN modified from AlexNet by stacking seven scene convolutional layers. Jin et al. \cite{JinEfficientDeepAesthetic2016} proposed ILGNet derived from part of the GoogLeNet which contains Inception module.
    
    \subsection{Deep Neural Networks for Computer Vision Tasks}
    {\bf Evolution of Architectures: }A variety of DNN models have been developed and achieved huge success in different computer vision tasks these years.
    AlexNet \cite{KrizhevskyImagenetclassificationdeep2012} was the first CNN to win the ImageNet Challenge in 2012 and it consists of five convolutional(CONV) layers and three fully-connected(FC) layers. VGG-16 \cite{SimonyanVeryDeepConvolutional2014} goes deeper to 16 layers consisting of 13 CONV layers and 3 FC layers. ResNet \cite{HeDeepResidualLearning2016} uses residual connections to go even deeper (34 layers or more). It was the first entry DNN in ImageNet Challenge that exceeded human-level accuracy with a top-5 error rate below 5\%.
    Previous work in \cite{KornblithBetterImageNetModels2018, RazavianCNNFeaturesofftheshelf2014} shows that, generally, the better performance a DNN could achieve in ImageNet classification task, the more effective deep features it could extract.
    
    \noindent{\bf DNN for Scene Recognition: }Scene recognition is a challenging problem since scenes not only provide visual information from the level of objects but also the relationship between them. Deep convolutional neural networks trained on places (Places-CNNs) have shown impressive results in scene recognition tasks \cite{ZhouPlaces10million2017,DonahueDeCAFDeepConvolutional2014} and have been applied in many areas.
    The content and scenery of an image are fundamental to its aesthetics and are sometimes overlooked in the assessment. Also, it's been proved that taking image content into account can improve the accuracy of image aesthetics prediction \cite{MaiCompositionPreservingDeepPhoto2016,Wangmultiscenedeeplearning2016}.
    
    \section{Method}
    \label{sec:method}
    In this section, we will give a detailed description of our method. As shown in Figure 2, we exploit three parallel deep neural networks and each of them is used to extract specific deep features from the input image. By aggregating the extracted features, we are able to classify them with a classifier.
    
    \subsection{Off-the-shelf CNN Features}
    Deep convolutional neural networks trained on a 1.2 million subset of the ImageNet dataset can be employed as a general feature extractor. Following the previous work \cite{RazavianCNNFeaturesofftheshelf2014, DonahueDeCAFDeepConvolutional2014}, we directly take the trained neural network weights from their original published work with no modification.
    
    DNNs like AlexNet and VGG-16 contain CONV layers at the top and FC layers at the bottom. We then directly take the 4096-dimensional activations from the first FC layer as the features that will be used later for classification. For ResNet, we need to utilize the features from its penultimate layer, i.e., the average pooling layer (AvgPool), which typically is 2048-dimensional.
    
    A recent work \cite{KornblithBetterImageNetModels2018} studies the effectiveness of ImageNet features and concludes that ResNet models are better extractors. We did similar experiments and the results are basically consistent. Thus, in our work, we choose ResNet-50 as the extractor and prove it's a better model on aesthetics assessment task by comparing it with AlexNet and VGG-16.

    \subsection{Using Composite Features From Different Nets}
    The advantage of our method is that we exploit three parallel deep neural networks to extract unique features from three different aspects, including the global view, local view and scene-aware information. We refer to the final aggregated features as {\it composite features} that will be used by the following classifier.
    
    \noindent{\bf Global View: }In order to extract effective features representing the picture as a whole, a column of DNN for image classification is used as the global view feature extractor. By resizing the given image to a fixed size and feeding it to the network, the global view features of this image could be acquired.
    
    \noindent{\bf Local View: }The local view of an image is closely related to its aesthetic evaluation and is often overlooked. Instead of using randomly sampling parts from the original high-resolution images, we crop the center area of the image by a fixed ratio (0.62) as the local view which is more visually representative since people pay more attention to the center. We deliver the cropped part to the identical deep neural network that has been previously used for global view and get the required feature vector.
    
    \noindent{\bf Scene-aware Information: }Besides global and local view, the content of an image has much to do with its overall aesthetics. Zhou et el. \cite{ZhouPlaces10million2017} published Places Database comprising 10 million scene photographs, labeled with 434 scene semantic categories. Among the DNNs they have trained, Places365-ResNet reaches 85.07\% top-5 accuracy, which is the highest of all. We then utilize the scene-aware features extracted by the Places365 model based on ResNet-50 as additional information to improve the performance of our method.
    
    \section{Experiments}
    We conducted experiments to evaluate the performance of different architectures of deep neural networks. We also analyze the contribution of local view and scene-aware information respectively. 
    
    \subsection{Datasets}
    AVA \cite{MurrayAVAlargescaledatabase2012} and CUHKPQ \cite{TangContentBasedPhotoQuality2013} are the datasets we use in our experiments. We build the subset AVA1 following \cite{MurrayAVAlargescaledatabase2012,JinEfficientDeepAesthetic2016,MaiCompositionPreservingDeepPhoto2016,Wangmultiscenedeeplearning2016}, and AVA2 following \cite{DongPhotoQualityAssessment2015,Wangmultiscenedeeplearning2016,WangFinetuningConvolutionalNeural2016}. And CUHKPQ is set up as \cite{WangFinetuningConvolutionalNeural2016}. Details are presented in Table \ref{table:expsetting}.

    \begin{table}[h!]
    \centering
        \begin{tabular}{c|c|c|c|c|l}
            \hline
            \multicolumn{2}{c|}{\bf Dataset} & {\bf High} & {\bf Low} & {\bf Train} & {\bf Test} \\
            \hline
            \multirow{2}{*}{\bf AVA} & {\bf AVA1} & 74,673 & 180,856 & 235,599 & 19,930 \\
            \cline{2-6}
                                & {\bf AVA2} & 25,553 & 25,553 & 25,553 & 25,553 \\
            \hline
            \multicolumn{2}{c|}{\bf CUHKPQ}  & 10,524 & 19,166 & 14,845 & 14,845 \\
            \hline
        \end{tabular}
        \caption{Experimental settings for AVA and CUHKPQ, including the details of high and low aesthetic pictures and the partition of training and testing sets.}
        \label{table:expsetting}
    \end{table}

    \subsection{Evaluating the Impact of Network Architectures}
    
    We extract features from three different DNNs as Section 3.1 described. The extracted features can be easily separated into different categories by traditional machine learning classifiers such as Support Vector Machine (SVM), Random Forest, AdaBoost and so on. Decent results could be achieved by these simple classifiers. Here, we adopt SVM with RBF kernel as our classifier. It is worth noting that we never try to fine-tune the classifier parameters so that we could get the authentic validity of the features.
    
    To find out which off-the-shelf deep neural network trained on ImageNet could generate the most effective features for aesthetics task, we carry out a basic experiment on AVA2 subset using pre-trained networks to learn deep abstractions which are then be classified with a simple SVM.
    
    \begin{table}[h!]
    \centering
        \begin{tabular}{c|c|c|c}
            \hline
            {\bf Model} & {\bf AlexNet} 
            & {\bf VGG-16} & {\bf ResNet-50} \\
            \hline
            {\bf AVA2} & 53.2
            & 82.1 & 87.7 \\
            \hline
            {\bf CUHKPQ} & 73.5
            & 87.1 & 90.3 \\
            \hline
        \end{tabular}{}
        \caption{Baseline accuracy on AVA2 and CUHKPQ using different DNN models with SVM as the classifier.}
        \label{table:baseline}
    \end{table}
    
    As Table \ref{table:baseline} shows, the deep features obtained from these various architectures resulted in different accuracy. AlexNet performs the worst of three. VGG-16 follows with better results. ResNet-50 achieves the highest among the three on both datasets. 
    
    It's obvious that utilizing ResNet models could bring about more generic features that lead to better accuracy in image aesthetics assessment, which is consistent with the conclusion from \cite{KornblithBetterImageNetModels2018} and experiments of Figure 3 further prove this.
    
    \subsection{Evaluating the Benefits of Local and Scene-aware Information}
    Deep representation extracted by a single pre-trained model alone is not sufficient for getting more promising result. Here, we'd like to demonstrate the influential contributions of the local and scene-aware information of the input image. The local view features are generated by the same deep neural network previously used by the global view. Moreover, to testify the effectiveness of scene-aware information, we used the features generated from the Places365-ResNet from \cite{ZhouPlaces10million2017}, which is based on ResNet-50 and fine-tuned for scene classification. 
    
    \begin{figure}[htb]
    \begin{minipage}[b]{0.48\linewidth}
      \centering
      \centerline{\includegraphics[width=3.8cm]{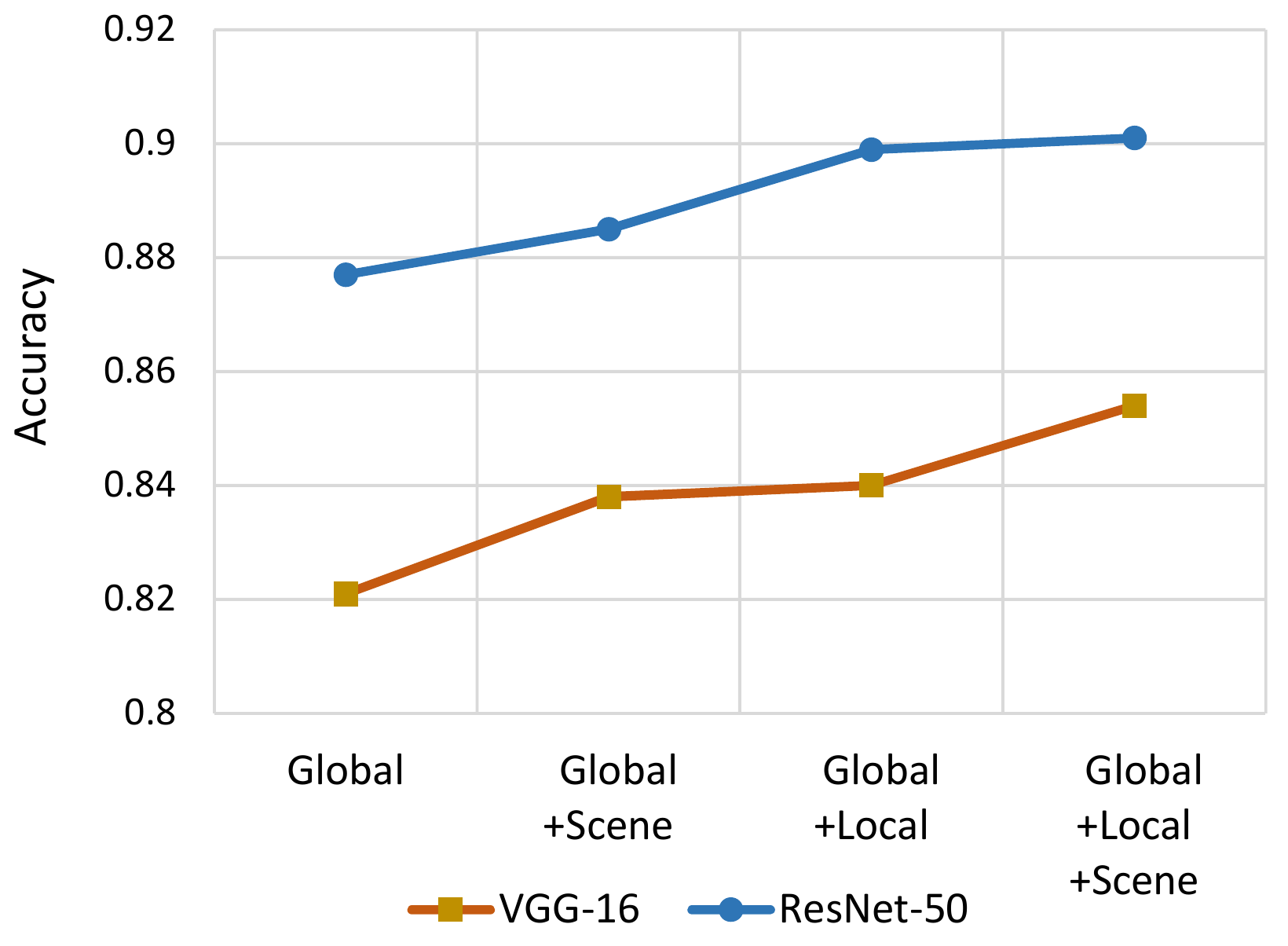}}
      \centerline{(a)}
      \medskip
    \end{minipage}
    \hfill
    \begin{minipage}[b]{0.48\linewidth}
        \centering
        \centerline{\includegraphics[width=3.8cm]{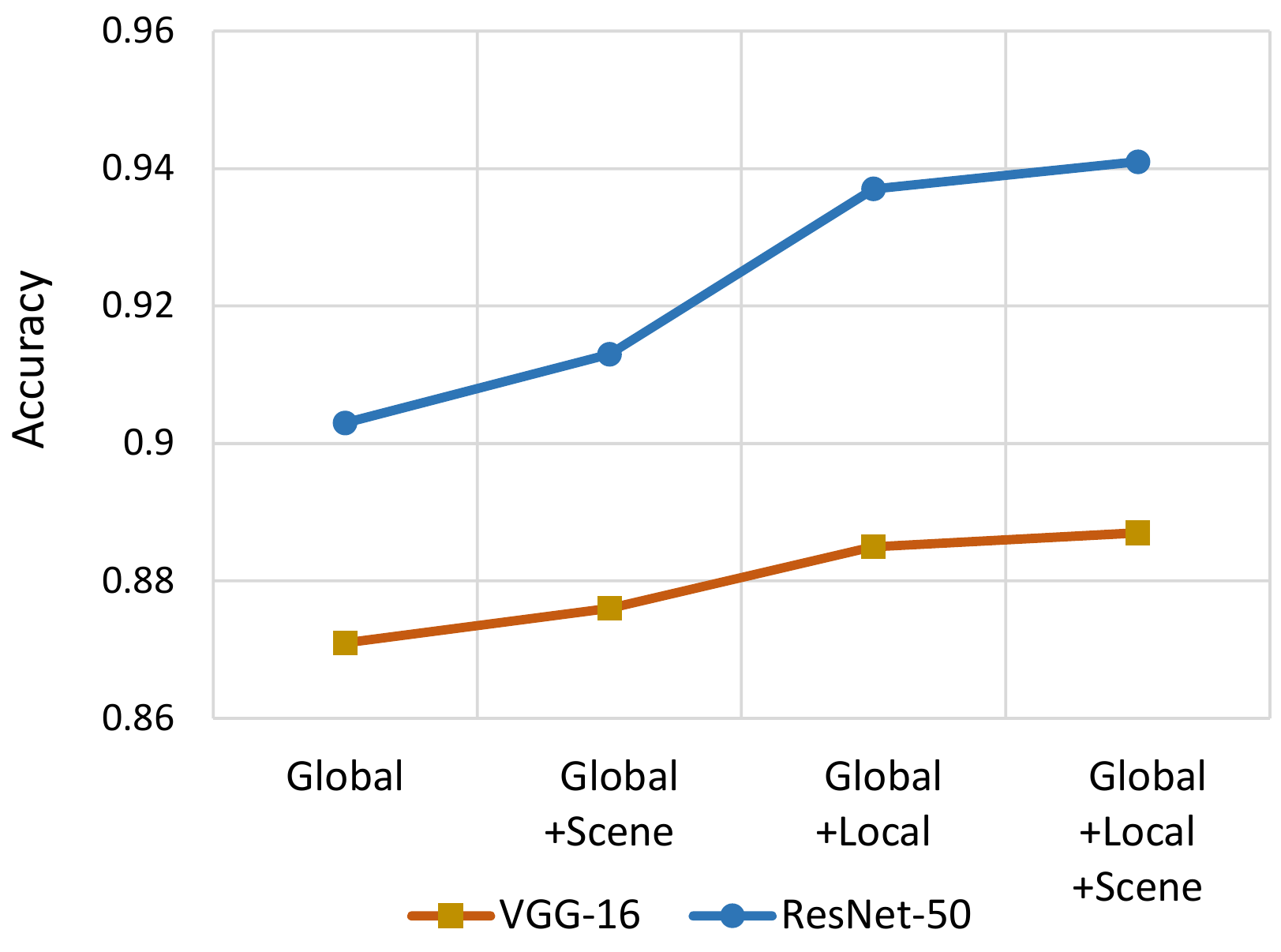}}
        \centerline{(b)}
        \medskip
    \end{minipage}

    \caption{Accuracy on two datasets for VGG-16 and ResNet-50 with different combinations of features. (a) results on AVA2, (b) results on CUHKPQ.}
    \label{fig:figure3}
    \end{figure}
    
    \begin{figure}[htb]
        \begin{minipage}[b]{1.0\linewidth}
          \centering
          \centerline{\includegraphics[width=4.0cm]{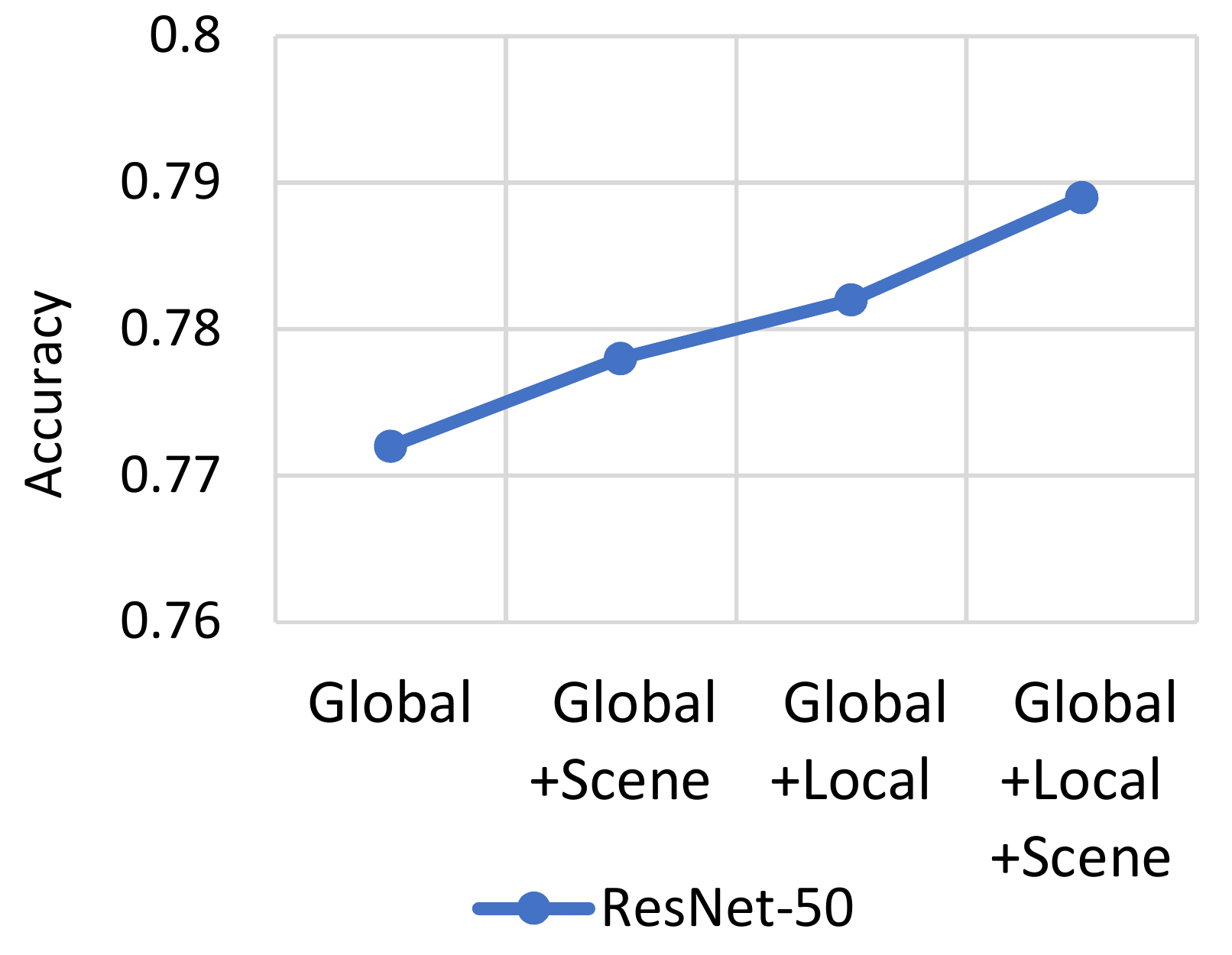}}\medskip
        \end{minipage}
        \caption{Accuracy on AVA1 subset using ResNet-50.}
        \label{fig:figure4}
    \end{figure}

    The benefits of local and scene-aware information are evaluated and presented in Figure 3. For AVA2 subset, with only global view, the accuracy of classifying features generated by VGG-16 is 82.1\% and by ResNet-50 is 87.7\%. With both global view and scene-aware information, the accuracy increases by 1.7\% and 0.8\% respectively. With global and local view, it rises by 1.9\% and 2.2\%. Finally, when we use the composite features made up of global, local and scene-aware information, the accuracy goes much higher and reaches 85.4\% and 90.0\% on AVA2 subset. 
    
    Further, we conduct similar experiment on a smaller dataset, CUHKPQ, and same tendency is found, as shown in Figure 3 (b). With the composite features from ResNet-50 used, our method achieves 94.1\%, which has already excelled the state-of-the-art result in \cite{WangFinetuningConvolutionalNeural2016}.
    
    According to Figure 3, ResNet-50 performs better, so we run experiments on AVA1 subset using ResNet-50. The increase of the accuracy is shown in Figure 4, regardless of the fact that the overall precision is restricted by the untuned SVM and the huge amount of pictures scoring around 5 which make them hardly separated. Again, the result shows the benefit of using composite features.
    
    Another experiment we do is cross-dataset evaluation. We train a SVM classifier based on the features extracted from the global view of the whole AVA2 dataset and use this model to classify the features from the entire CUHKPQ dataset. The result, 87.2\%, further verifies the effectiveness of both the deep representation gained from ResNet model and the SVM model.
    
    All these results together prove that by using more kinds of features extracted from the image, better performance could be acquired. Meanwhile, deep neural network like ResNet-50 is more excellent at extracting generic deep features containing aesthetics information than VGG-16 and AlexNet.
    
    \subsection{Comparison with the State-of-the-Art}

    Table \ref{table:compare} shows the results of our proposed method on AVA2 subset for image aesthetics categorization. It is obvious that our method achieves the state-of-the-art result compared to other recently proposed methods. Specifically, by using the composite features generated by pretrained ResNet-50 for image classification and scene recognition, we first bring the accuracy on AVA2 subset up to 90.01\%, which outperforms all the existing methods.

    \begin{table}[h!]
        \centering
        \begin{tabular}{c|c|c}
            \hline
            \thead{\bf Methods} & \thead{{\bf Train/Fine-tune} \\ {\bf Deep Models}}  & \thead{\bf Acc\\\bf(\%)} \\
            \hline \hline
            Tian et al. (2015) \cite{TianQueryDependentAestheticModel2015} & Yes & 80.38 \\
            \hline
            Dong et al. (2015) \cite{DongPhotoQualityAssessment2015} & No & 83.52 \\
            \hline
            Wang et al. (2016) \cite{Wangmultiscenedeeplearning2016} & Yes & 84.88 \\
            \hline
            Jin et al. (2017) \cite{JinEfficientDeepAesthetic2016} & Yes & 85.53 \\
            \hline \hline
            {\bf{VGG-16 \scriptsize{Global+Local+Scene}}} & No & {\bf 85.40}\\
            \hline
            {\bf ResNet-50 \scriptsize{Global+Local+Scene}} & No & {\bf 90.01}\\
            \hline
        \end{tabular}{}
        \caption{Comparisons with the state-of-the-art on AVA2.}
        \label{table:compare}
    \end{table}
    
    \section{Conclusion}
    This paper presents an effective and efficient scheme that using composite features generated from deep pretrained convolutional neural networks leads to an increase on the accuracy of image aesthetics assessment. Our proposed training-free method takes the local, global and scene-aware information of images into consideration and utilizes the off-the-shelf deep learning models in the procedure of feature extracting. Our experimental analysis demonstrates that our method achieves superior performance in comparison to other state-of-the-art approaches.

    \vfill
    \pagebreak

    \bibliographystyle{IEEEbib}
    \bibliography{ref}

\begin{thebibliography}{10}

\bibitem{DengImageAestheticAssessment2017}
Y.~Deng, C.~C. Loy, and X.~Tang,
\newblock ``Image {{Aesthetic Assessment}}: {{An}} experimental survey,''
\newblock {\em IEEE Signal Processing Magazine}, vol. 34, no. 4, pp. 80--106,
  July 2017.

\bibitem{MurrayAVAlargescaledatabase2012}
N.~Murray, L.~Marchesotti, and F.~Perronnin,
\newblock ``{{AVA}}: {{A}} large-scale database for aesthetic visual
  analysis,''
\newblock in {\em 2012 {{IEEE Conference}} on {{Computer Vision}} and {{Pattern
  Recognition}}}, June 2012, pp. 2408--2415.

\bibitem{KrizhevskyImagenetclassificationdeep2012}
Alex Krizhevsky, Ilya Sutskever, and Geoffrey~E. Hinton,
\newblock ``Imagenet classification with deep convolutional neural networks,''
\newblock in {\em Advances in {{Neural Information Processing Systems}}}, 2012,
  p. 2012.

\bibitem{SimonyanVeryDeepConvolutional2014}
Karen Simonyan and Andrew Zisserman,
\newblock ``Very {{Deep Convolutional Networks}} for {{Large}}-{{Scale Image
  Recognition}},''
\newblock {\em arXiv:1409.1556 [cs]}, Sept. 2014.

\bibitem{HeDeepResidualLearning2016}
K.~He, X.~Zhang, S.~Ren, and J.~Sun,
\newblock ``Deep {{Residual Learning}} for {{Image Recognition}},''
\newblock in {\em 2016 {{IEEE Conference}} on {{Computer Vision}} and {{Pattern
  Recognition}} ({{CVPR}})}, June 2016, pp. 770--778.

\bibitem{ZhouPlaces10million2017}
B.~Zhou, A.~Lapedriza, A.~Khosla, A.~Oliva, and A.~Torralba,
\newblock ``Places: {{A}} 10 million {{Image Database}} for {{Scene
  Recognition}},''
\newblock {\em IEEE Transactions on Pattern Analysis and Machine Intelligence},
  vol. PP, no. 99, pp. 1--1, 2017.

\bibitem{LuRAPIDRatingPictorial2014}
Xin Lu, Zhe Lin, Hailin Jin, Jianchao Yang, and James~Z. Wang,
\newblock ``{{RAPID}}: {{Rating Pictorial Aesthetics}} using {{Deep
  Learning}},''
\newblock in {\em Proceedings of the 22nd {{ACM}} International Conference on
  {{Multimedia}}}. 2014, pp. 457--466, {ACM}.

\bibitem{DengImageNetlargescalehierarchical2009}
J.~Deng, W.~Dong, R.~Socher, L.~J. Li, Kai Li, and Li~Fei-Fei,
\newblock ``{{ImageNet}}: {{A}} large-scale hierarchical image database,''
\newblock in {\em 2009 {{IEEE Conference}} on {{Computer Vision}} and {{Pattern
  Recognition}}}, June 2009, pp. 248--255.

\bibitem{RazavianCNNFeaturesofftheshelf2014}
Ali~Sharif Razavian, Hossein Azizpour, Josephine Sullivan, and Stefan Carlsson,
\newblock ``{{CNN Features}} off-the-shelf: An {{Astounding Baseline}} for
  {{Recognition}},''
\newblock {\em arXiv:1403.6382 [cs]}, Mar. 2014.

\bibitem{DonahueDeCAFDeepConvolutional2014}
Jeff Donahue, Yangqing Jia, Oriol Vinyals, Judy Hoffman, Ning Zhang, Eric
  Tzeng, and Trevor Darrell,
\newblock ``{{DeCAF}}: {{A Deep Convolutional Activation Feature}} for
  {{Generic Visual Recognition}},''
\newblock in {\em Proceedings of the 31st {{International Conference}} on
  {{International Conference}} on {{Machine Learning}} - {{Volume}} 32},
  Beijing, China, 2014, ICML'14, pp. I--647--I--655, {JMLR.org}.

\bibitem{JinEfficientDeepAesthetic2016}
Xin Jin, Le~Wu, Zheyuan He, Siyu Chen, Jingying Chi, Siwei Peng, Xiaodong Li,
  and Shiming Ge,
\newblock ``Efficient {{Deep Aesthetic Image Classification}} using {{Connected
  Local}} and {{Global Features}},''
\newblock {\em arXiv:1610.02256 [cs]}, Oct. 2016.

\bibitem{DattaStudyingAestheticsPhotographic2006}
Ritendra Datta, Dhiraj Joshi, Jia Li, and James~Z. Wang,
\newblock ``Studying {{Aesthetics}} in {{Photographic Images Using}} a
  {{Computational Approach}},''
\newblock in {\em Computer {{Vision}} \textendash{} {{ECCV}} 2006}. May 2006,
  Lecture Notes in Computer Science, pp. 288--301, {Springer, Berlin,
  Heidelberg}.

\bibitem{MarchesottiDiscoveringBeautifulAttributes2015}
Luca Marchesotti, Naila Murray, and Florent Perronnin,
\newblock ``Discovering {{Beautiful Attributes}} for {{Aesthetic Image
  Analysis}},''
\newblock {\em International Journal of Computer Vision}, vol. 113, no. 3, pp.
  246--266, July 2015.

\bibitem{MaiCompositionPreservingDeepPhoto2016}
L.~Mai, H.~Jin, and F.~Liu,
\newblock ``Composition-{{Preserving Deep Photo Aesthetics Assessment}},''
\newblock in {\em 2016 {{IEEE Conference}} on {{Computer Vision}} and {{Pattern
  Recognition}} ({{CVPR}})}, June 2016, pp. 497--506.

\bibitem{TianQueryDependentAestheticModel2015}
X.~Tian, Z.~Dong, K.~Yang, and T.~Mei,
\newblock ``Query-{{Dependent Aesthetic Model With Deep Learning}} for {{Photo
  Quality Assessment}},''
\newblock {\em IEEE Transactions on Multimedia}, vol. 17, no. 11, pp.
  2035--2048, Nov. 2015.

\bibitem{DongPhotoQualityAssessment2015}
Zhe Dong, Xu~Shen, Houqiang Li, and Xinmei Tian,
\newblock ``Photo {{Quality Assessment}} with {{DCNN}} that {{Understands Image
  Well}},''
\newblock in {\em {{MultiMedia Modeling}}}. Jan. 2015, Lecture Notes in
  Computer Science, pp. 524--535, {Springer, Cham}.

\bibitem{Wangmultiscenedeeplearning2016}
Weining Wang, Mingquan Zhao, Li~Wang, Jiexiong Huang, Chengjia Cai, and
  Xiangmin Xu,
\newblock ``A multi-scene deep learning model for image aesthetic evaluation,''
\newblock {\em Signal Processing: Image Communication}, vol. 47, no. Supplement
  C, pp. 511--518, Sept. 2016.

\bibitem{KornblithBetterImageNetModels2018}
Simon Kornblith, Jonathon Shlens, and Quoc~V. Le,
\newblock ``Do {{Better ImageNet Models Transfer Better}}?,''
\newblock {\em arXiv:1805.08974 [cs, stat]}, May 2018.

\bibitem{TangContentBasedPhotoQuality2013}
X.~Tang, W.~Luo, and X.~Wang,
\newblock ``Content-{{Based Photo Quality Assessment}},''
\newblock {\em IEEE Transactions on Multimedia}, vol. 15, no. 8, pp.
  1930--1943, Dec. 2013.

\bibitem{WangFinetuningConvolutionalNeural2016}
Yeqing Wang, Yi~Li, and F.~Porikli,
\newblock ``Finetuning {{Convolutional Neural Networks}} for visual
  aesthetics,''
\newblock in {\em 2016 23rd {{International Conference}} on {{Pattern
  Recognition}} ({{ICPR}})}, Dec. 2016, pp. 3554--3559.

\end{thebibliography}
    
    \end{document}